%% file: root.tex
\documentclass[letterpaper, 10 pt, conference]{ieeeconf}  

\IEEEoverridecommandlockouts                              

\makeatletter
\let\NAT@parse\undefined
\makeatother
\usepackage[numbers]{natbib}

\usepackage[pdftex]{graphicx}
\usepackage{amsmath}
\usepackage{amssymb}  
\usepackage{subfig}
\usepackage{multirow}
\usepackage{array,booktabs}
\usepackage{diagbox}
\usepackage{balance}
\usepackage[ruled]{algorithm}
\usepackage{algpseudocode}
\usepackage{bm}

 \usepackage[final]{hyperref}
 \hypersetup{
     colorlinks=true,
     linkcolor=blue,
     filecolor=magenta,
     urlcolor=blue,
     citecolor=black
 }
\usepackage[font=small]{caption}
\usepackage{./rpm_packages/rpm_SIunits}
\usepackage{./rpm_packages/rpm_acronyms}
\usepackage{./rpm_packages/rpm_math}
\usepackage{./rpm_packages/rpm_misc}

\usepackage{soul,color}
\usepackage{lipsum}

\usepackage{xcolor}

\usepackage{tikz} 

\title{\LARGE \bf
2D Ego-Motion with Yaw Estimation using Only \\ mmWave Radars via Two-Way weighted ICP
}     

\author{Hojune Kim, Hyesu Jang,  and Ayoung Kim${}^{*}$
\thanks{H. Kim is with the Department of Aerospace Engineering, SNU, Seoul, S. Korea. H. Jang, and A. Kim are with the Department of Mechanical Engineering, SNU, Seoul, S. Korea {\tt\small [hojjunekim, dortz, ayoungk]@snu.ac.kr}}%
}

\begin{document}

\maketitle
\thispagestyle{empty}
\pagestyle{empty}

\input{abstract}
\input{introduction}
\input{relatedwork}

\input{method}

\input{experiment}
\input{conclusion}


\balance
\small
\bibliographystyle{IEEEtranN} 
\bibliography{string-short,references}

\end{document}

%% file: abstract.tex
\begin{abstract}
The interest in single-chip mmWave Radar is driven by their compact form factor, cost-effectiveness, and robustness under harsh environmental conditions. Despite its promising attributes, the principal limitation of mmWave radar lies in its capacity for autonomous yaw rate estimation. Conventional solutions have often resorted to integrating \ac{IMU} or deploying multiple radar units to circumvent this shortcoming. This paper introduces an innovative methodology for two-dimensional ego-motion estimation, focusing on yaw rate deduction, utilizing solely mmWave radar sensors. By applying a weighted \ac{ICP} algorithm to register processed points derived from heatmap data, our method facilitates 2D ego-motion estimation devoid of prior information. Through experimental validation, we verified the effectiveness and promise of our technique for ego-motion estimation using exclusively radar data.


\end{abstract}

%% file: introduction.tex
\section{Introduction}
\label{sec:intro}

Radar technology emerges as a promising solution for perception in robotics and autonomous systems due to its potential to overcome the limitations of traditional sensors like LiDAR and cameras. LiDAR systems are often hampered by their substantial size and associated costs, and cameras struggle with fluctuations in lighting conditions. In contrast, millimeter-wave radar emerges as a compact and economically feasible alternative, distinguished by its proficiency in estimating ego-motion within robotic frameworks in various environments.
Despite the potential of radar, challenges persist in fully exploiting its capabilities. mmWave radar generates cube type data via the \ac{FFT} of chirp signals. Although there has been notable advancement in processing these signals for analysis, a discernible gap exists in the methodologies to leverage the lower-level radar data.
Current approaches frequently employ radar in synergy with additional sensors, such as \ac{IMU} or LiDAR systems, to estimate ego-motion.\cite{zhou2022towards} This strategy primarily stems from the challenges associated with processing the high-level 4D point cloud outputs derived from radar data.

This paper proposes a novel Two-Way weighted \ac{ICP} approach exclusively using mmWave radar for 2D ego-motion estimation, with a specific emphasis on yaw estimation. By eliminating the need for additional sensors, this research aims to overcome the limitations associated with traditional sensor fusion approaches.
Our approach aims to enhance perception capabilities, offering advantages in cost-effectiveness, compactness, and resilience to environmental conditions. Through verification with publicly available datasets, we demonstrate the efficacy and potential of radar for perception tasks in real-world environments.

\begin{figure}[!t]
    \centering
    \includegraphics[width=\columnwidth]{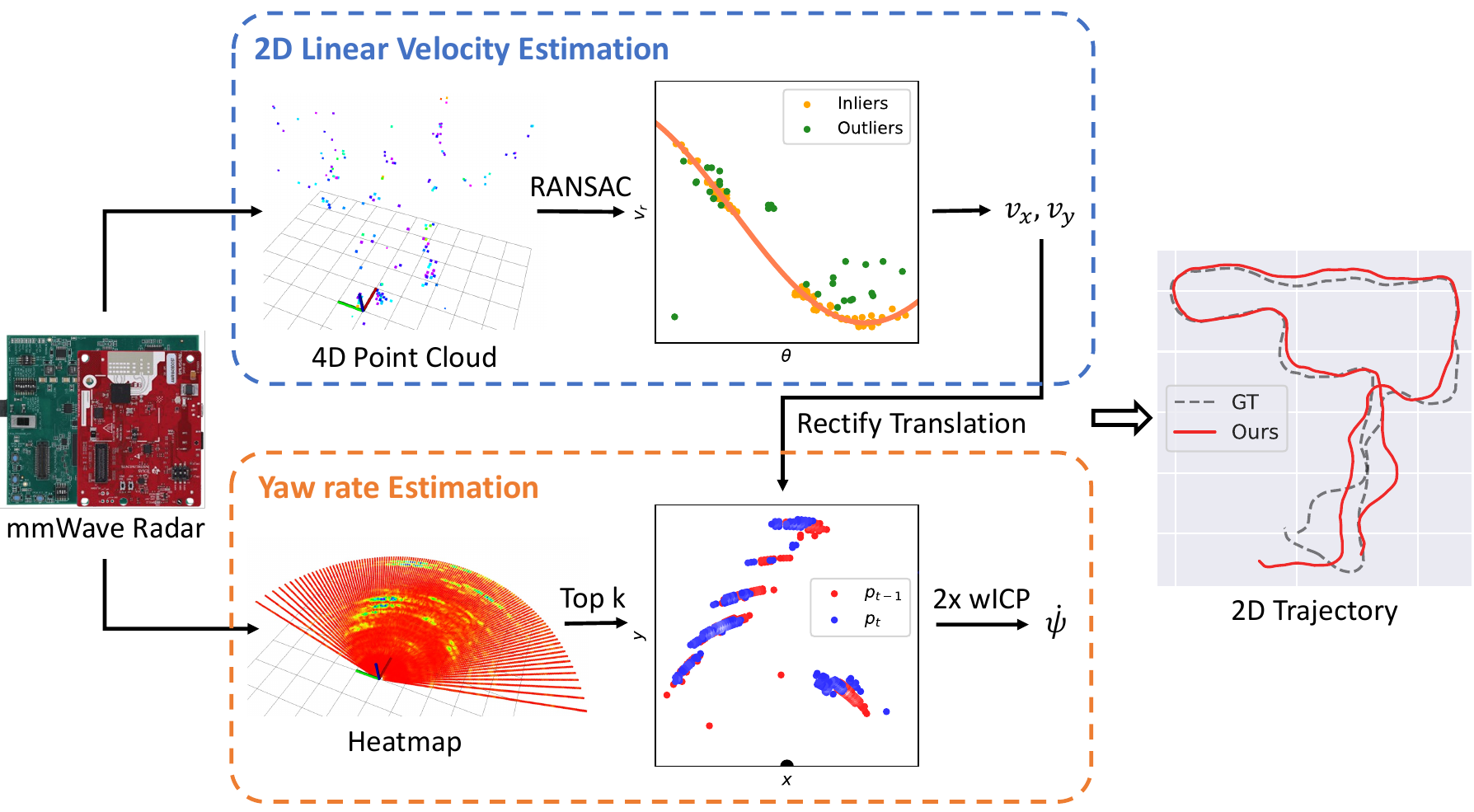}
    \caption{Visualization of the 2D ego-motion estimation process using only mmWave radar, along with the estimated trajectory in an indoor environment.}
    \label{fig:intro}
    \vspace{-5mm}
\end{figure}

\begin{itemize}
    \item \textbf{Non-learning radar-sole 2D ego-motion estimation}
    This paper presents the 
    implementation of 2D ego-motion estimation, including rotation, solely utilizing mmWave radar data without integrating other sensors or needing a GPU.
    
    \item \textbf{Clutter matching via feature sampling and two-way weighted \ac{ICP}} 
    We effectively managed to match clutter mmWave radar data by employing feature sampling and a two-way weighted \ac{ICP} approach.

    \item \textbf{Validation of radar-only planar odometry}
    The validity of our pipeline was verified through radar-only planar odometry performed on a public dataset.
\end{itemize}

%% file: relatedwork.tex
\section{related work}
\label{sec:relatedwork}
\begin{figure*}[!t]
    \centering
    \includegraphics[width=0.9\textwidth]{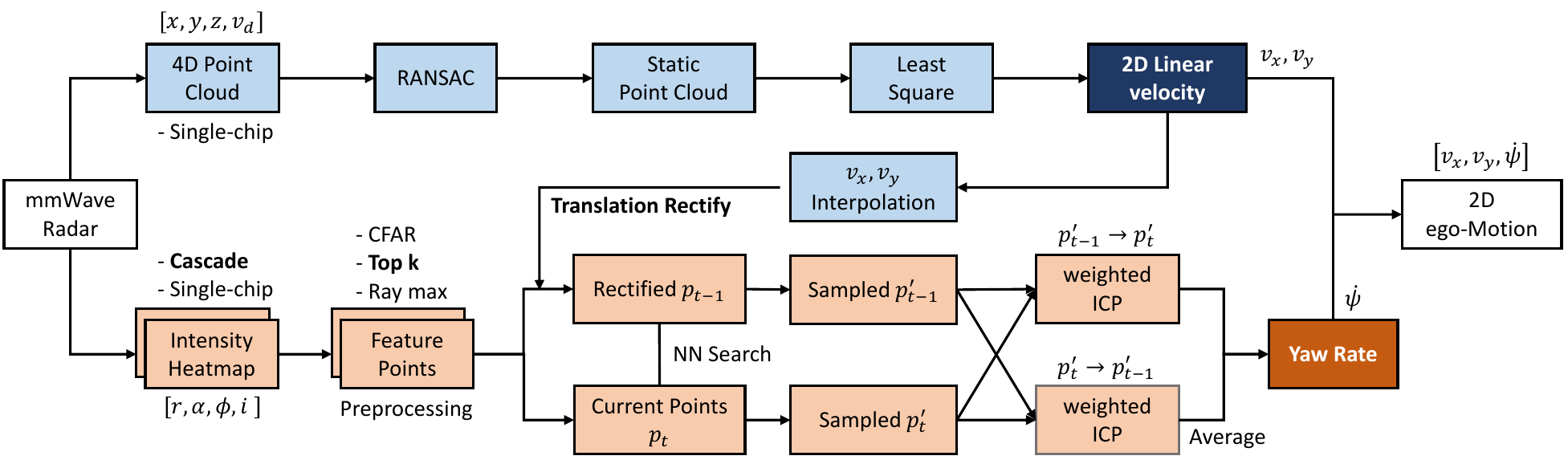}
    \caption{Overview of the proposed method to estimate 2D ego-motion divided by two sections.}
    \label{fig:pipeline}
    \vspace{-5mm}
\end{figure*}

There are two type of mmWave radar, mechanically scanning radar and single-chip radar. Scanning radar provide high angular resolution images, facilitating point matching and widely applied in odometry or place recognition. However, bulky size, high-cost and high-power consumption makes unsuitable to the small mobile robots. Conversely, single-chip radar boasts a low cost and compact design, garnering increasing interest recently. Single-chip radar offers 4D point clouds (range, azimuth, elevation, and Doppler velocity), extracted from the radar data cube through signal processing, as well as heatmaps generated by applying \ac{FFT} to the data cube. We analyzed methods utilizing single-chip radar, particularly focusing on ego-motion estimation.

\subsection{4D Point Cloud based Ego-Motion Estimation}
Most of the studies to estimate ego-motion by using radar utilized point clouds with doppler velocity from radar. First, by using single-chip radar solely, the linear velocity can be estimated by analyzing doppler velocities over azimuth angles.\cite{kellner2014instantaneous} However, since the doppler velcities are radial direction, to estimate yaw rate as well, using multiple radar sensors in \cite{park20213d}, or assuming an Ackerman model are proposed in \cite{kellner2014instantaneous}. In the other side, inertial sensors are usually combined with radar to enhance the performance. In \cite{doer2020ekf} proposed Extended Kalman Filter to use IMU and ego-velocity from radar, and \cite{lu2020milliego}\cite{almalioglu2020milli} use deep learning networks by combining IMU.

\subsection{Heatmap based Ego-Motion Estimation}
Relatively, there are few heatmap based research, the interest become rising recently. Due to the large number of unprocessed points in the heatmap, learning-based ego-motion estimation is a commom method. In \cite{rai20234dego}, a supervised learning method was trained to estimate ego-motion by 3D intensity heatmap with IMU data, but its accuracy for the trajectory has not been verified. In \cite{zhao20213d}, a deep network was trained to estimate the position directly from the heatmap. 
However, the development of heatmap-based method has mainly been focused on learning-based approaches, rather than being addressed by intuitively matching techniques.


%% file: method.tex
\section{Method}
\label{sec:method}

To utilize the information from the low-level heatmap of the mmWave radar, we employed both single-chip radar and cascade radar to estimate the 2D ego-motion. While both configurations yield a heatmap, the cascade radar is preferred for estimating yaw rotation because it can generate a comprehensive range-elevation-azimuth intensity heatmap. Conversely, the single-chip radar, characterized by its 4D point cloud encompassing Doppler velocity, is utilized to estimate 2D linear velocity. Consequently, the methodology for estimating 2D ego-motion was structured into two distinct phases: linear velocity and yaw rate estimation, as depicted in \figref{fig:pipeline}.
\subsection{2D Linear Velocity Estimation}
First, we estimated the 2D linear velocity with the doppler velocity obtained from targets detected by single-chip radar, which introduced in \cite{kellner2014instantaneous}. Since the doppler velocity represents the velocity of objects relative to the radar, the radial velocity with respect to the azimuth $\theta$ follows the sinusoidal curve:
\begin{equation}
\begin{bmatrix}
v_{r,1} \\
\vdots \\
v_{r,n}
\end{bmatrix} = \begin{bmatrix}
\cos\theta_1 & \sin\theta_1 \\
\vdots & \vdots \\
\cos\theta_n & \sin\theta_n
\end{bmatrix} \begin{bmatrix}
v_x \\
v_y
\end{bmatrix}
\end{equation}
with $v_{r} = \frac{\sqrt{x^2+y^2}}{r}v_d$, the radial velocity projected to the xy plane.

Generally, most points detected are static, facilitating the robust estimation of 2D linear velocity by applying a sine curve fitting to the data utilizing the \ac{RANSAC}. Upon identifying inlier static objects, the computation of 2D linear velocity is refined to a least squares problem. When the number of inliers exceeds three, it becomes feasible to calculate a 3D linear velocity. However, due to the relative inaccuracy of height measurements, the proposed algorithm is confined to linear velocities along the x and y axes.

\subsection{mmWave Heatmap Preprocessing}
Several methods exist for estimating rotation from the 2D intensity map, including the \ac{ICP}, a classical method, and the power-azimuth method, frequently employed in mechanically spinning radars. However, given the mmWave radar's limited detection range and irregular distribution across azimuth and elevation, the classical ICP method is suitable for use with mmWave radar. Since the heatmap from the cascade radar constitutes an unfiltered fixed point cloud involving over one million points, the initial feature extraction step necessitates simplicity. First, \ac{CFAR} is a common preprocessing approach in radar \cite{hm1968adaptive}, that dynamically adjusts thresholds based on local noise levels. Second, \textbf{Top-k} points with the highest intensity values as features, which serves as a straightforward method for feature extraction. Lastly, with the heatmap being segmented by azimuth, it permits conceptualization as a 2D LiDAR. Each ray is represented by its point of highest intensity, which are referred to as \textbf{Ray-max}. To the best of our knowledge, there has yet to be a successful application of feature point registration from the heatmap in single-chip radar for yaw estimation. Therefore, we compared these preprocessing techniques to solve the 2D registration challenge.

The timestamps of single-chip radar and cascade radar are different, we interpolated the linear velocity between the timestamps of single-chip radar and cascade radar data to ensure synchronization in motion estimation. When the heatmap is derived from single-chip radar data, the frame rate is same with the 4D point cloud as $dt_s=10$Hz. Conversely, when employing data from the cascade radar system, the frame rate is reduced to $dt_c=5$Hz. Under these conditions, we linearly interpolated the linear velocities which achieved between two successive cascade radar frames. This interpolated velocity is then utilized to correct translational errors in the heatmap.
\begin{equation}
\hat{\bm{v}}_c = \frac{\hat{\bm{v}}_{s_t}-\hat{\bm{v}}_{s_{t-1}}}{dt_s} \left( \frac{t_{c_t}+t_{c_{t-1}}}{2}-t_{s_{t-1}} \right) + \hat{\bm{v}}_{s_{t-1}}
\end{equation}
\begin{equation}
\bm{p}_{t-1}^{rec} = \bm{p}_{t-1} - \hat{\bm{v}}_c dt_c
\end{equation}
The subscript $s$ denotes single-chip radar and $c$ represents cascade radar. Consequently, we obtain rectified feature points $\bm{p}_{t-1}^{rec}$.

Despite only the rotational difference remaining between $\bm{p}_{t-1}^{rec}$ and $\bm{p}_t$, establishing an initial guess devoid of the \ac{IMU} data and achieving a converged solution presents a significant challenge. Therefore, to attain a converged solution within a limited number of iteration, we set an error function as a weighted sum of the errors in range and azimuth:
\begin{equation}
e_{ij} = \alpha(r_i-r_j)^2+\beta(\theta_i-\theta_j)^2
\end{equation}
To ensure accurate source-target correspondence, we establish two threshold parameters, $\epsilon_{max}$ and $\epsilon_{min}$, which determine the criteria for categorizing source points based on their proximity to target points. Source points are categorized as \textit{remove} or \textit{neglect} if they are respectively too far or too close to any target point; those remaining are considered potential matches with the closest target point, facilitating \ac{ICP} computations. \textit{remove} points are discarded for the iteration, while \textit{neglect} points are retained for subsequent iterations. Additionally, points from the rectified previous iteration, may move beyond the radar's observation range, designating points outside the \ac{ROI} as \textit{neglect} points.
\begin{equation}
\bm{p}_{src} =
\begin{cases}
    \textit{remove} & \text{if } e_{min} > \epsilon_{max} \\
    \textit{neglect} & \text{if } e_{min} < \epsilon_{min} \text{ or out of \ac{ROI}} \\
    \textit{match} & \text{else}
\end{cases}
\end{equation}
This sampling method can lead to the optimal solution faster, even avoid local minima because the \textit{neglect} points can be used as matched points in the next iteration step. \\

\begin{figure}[!t]
    \centering
    \includegraphics[width=0.95\linewidth]{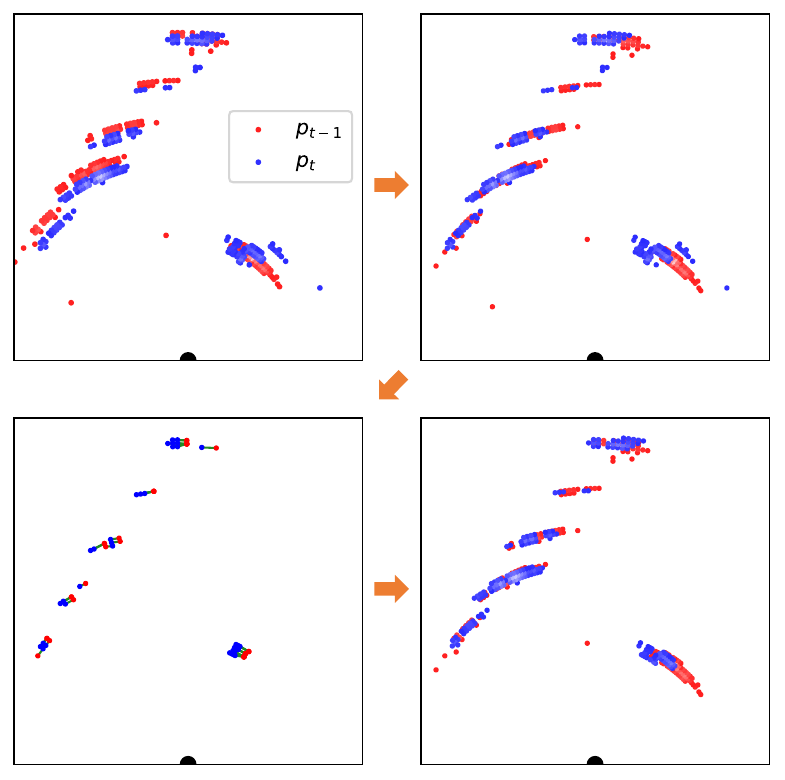}
    \caption{Illustration of our key point matching step. From the intensity heatmap, preprocess the Top k points(top left) and rectify the previous frame points by estimated linear velocities(top right). After sampling the points, find the matched pair to perform weighted \ac{ICP}(bottom left). Aligned points at 1st iteration in current source to previous target(bottom right). Black point on the bottom is the origin which represents the robot position.}
    \label{fig:icp}
\end{figure}

\begin{figure*}[!t]
    \centering
    \includegraphics[width=0.95\textwidth]{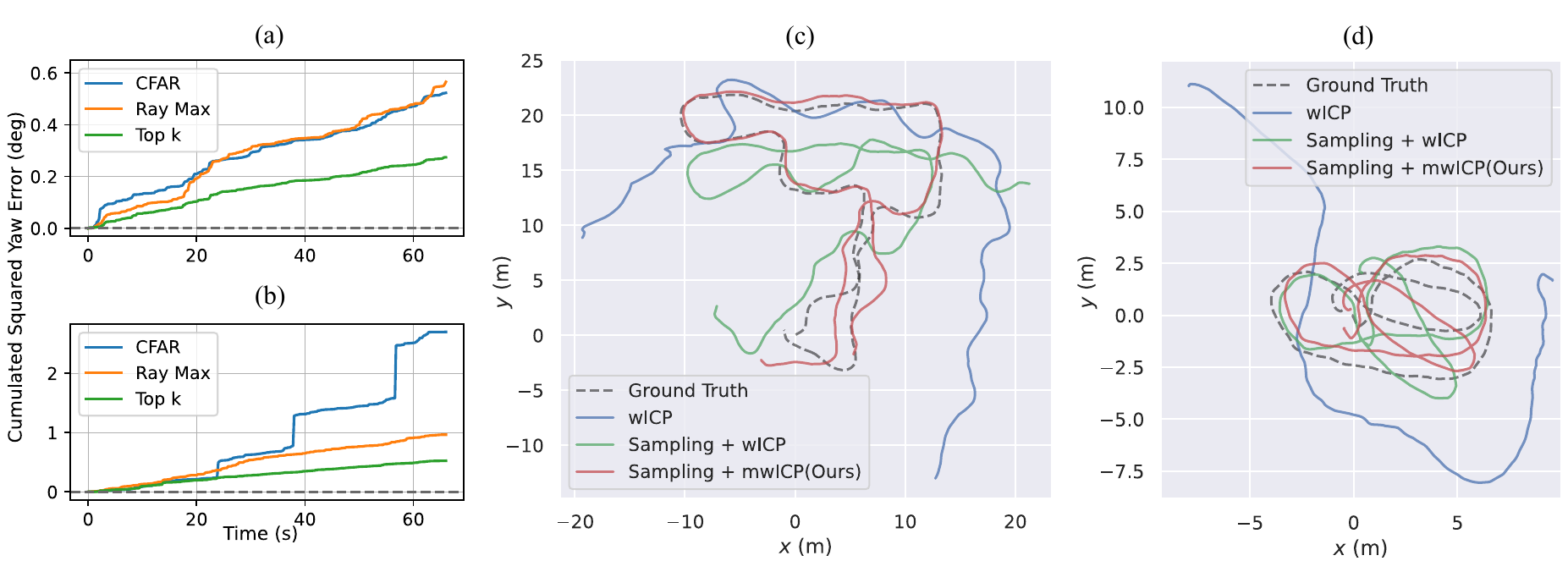}
    \caption{Cumulative squared error of the estimated yaw in (a)EC Hallways 0 sequence and (b)Aspen 5 sequence respect to the preprocessing methods. Through the estimated 2D ego-motion,  the trajectories of each sequences (c) and (d) with Top k preprocessing method. Preprocessing with one-way weighted \ac{ICP} represents using the previous frame as the source and the current frame as the target.}
    \label{fig:result}
    \vspace{-5mm}
\end{figure*}

\subsection{Weighted Iterative Closest Point}
Leveraging the intensity information inherent to each sampled points, we can implement a weighted \ac{ICP} algorithm. Since the intensity values within a heatmap can fluctuate even in a successive data, directly incorporating these values into the error function proves challenging. However, their utility as weights is beneficial. The correspondence between source and target is defined by error function, and the translation has been solved by estimated linear velocities. We applied weighted \ac{ICP}, as described in \cite{bergstrom2014robust}, which utilizes normalized intensity information as weights solely for rotational registration. While denoting $\bm{x}_i$ and $\bm{y}_i$ as the source and target correspondences, it derived as
\begin{equation}
\begin{aligned}
R^* &= \underset{R}{\arg\min} \sum_{i} w_i \| \bm{y}_i - R\bm{x}_i \|^2 \\
    &= \underset{R}{\arg\max}\sum_{i} w_i\bm{y}_i^TR\bm{x}_i = \underset{R}{\arg\max} \, \text{tr}(R^TH) \\
H &= \sum_{i}w_i\bm{x}_i\bm{y}_i^T = UDV^T, \ \ R^* = UV^T \in SO(2)
\end{aligned}
\end{equation}
which the optimal rotation matrix obtained by singular value decomposition of matrix $H$. The Rotation matrix $R$ only contains yaw rotation and the weight is a normalized intensity in our case.

The process of fitting heatmap feature points to others is shown in \figref{fig:icp}. 
It is observed that errors in translation are effectively mitigated through the implementation of a 2D linear velocity estimation module. Additionally, the registration of sampled points demonstrates satisfactory performance, even within a single iteration. Despite the ability of the one-directional \ac{ICP} algorithm to achieve convergence, it is important to note that this does not assure the optimal solution. Unlike other point cloud registration problems, the segment size of the radar feature points is not maintained and mat not persist in subsequent frames. This indicates that the results may differ based on the designated source and target. Thus, to enhance the precision in estimating yaw rates, a weighted \ac{ICP} is executed bidirectionally and provides a mean rotation.


%% file: experiment.tex
\section{experiment}
\label{sec:experiment}

\subsection{Experiment Environment}
To verify the reliability of our method, we used ColoRadar dataset \cite{kramer2022coloradar} which provides the heatmap and 4D point clouds from single-chip and cascade mmWave radars. They used Texas Instruments MMWCAS-RF-EVM cascade radar, AWR1843BOOST-EVM single chip radar and DCA1000-EVM for raw data capture. As we are estimating 2D linear velocity and rotation without considering vertical, roll, or pitch movements, we selected two sequences suitable for verifying our method in the ColoRadar dataset.

\begin{table}[h!]
\centering
\begin{tabular}{c|ccc}
\textbf{Dataset}                    & \textbf{Distance} & \textbf{Time} & \textbf{Environments} \\ \hline\hline
EC Hallway 0       & 106.3m             & 98.8s                        & Hallway            \\ 
Aspen 5   & 46.7m             & 65.6s                        & Room         \\ 
\end{tabular}
\caption{Dataset Attributes}\label{tab:dataset}
\vspace{-3mm}
\end{table}

\begin{table}[h!]
\centering
\begin{tabular}{c|ccccc}
\textbf{Sensor}            & \textbf{Range} &  \textbf{Max.}  &  \textbf{Max.} & \textbf{Framerate} \\ 

  &  \textbf{resolution}  & \textbf{Range}  & \textbf{Azimuth}   &  \\ \hline\hline
Cascade       & 0.06m             & 7.6m   &   76.3$\degree$                        & 5Hz        \\ 
Single-chip   & 0.125m            & 8.0m   &   78.3$\degree$                        & 10Hz         \\ 
\end{tabular}
\caption{mmWave Radar Attributes}\label{tab:sensor}
\vspace{-3mm}
\end{table}

\subsection{Evaluation Criteria}
\subsubsection{Preprocessing Evaluation}
Three preprocessing methods were applied to estimate yaw rate using a two-way weighted \ac{ICP}. These approaches leverage identical inputs, heatmap and estimated 2D linear velocity. Therefore, the most suitable metric for assessing the efficacy of these preprocessing techniques is the evaluation of the estimated yaw error. The entire pipeline for estimating 2D ego motion proceeds independently without prior knowledge of previous frames. Thus, the assessment of these methods is conducted through the analysis of the accumulated squared error of the estimated yaw.
\subsubsection{2D Trajectory}
Upon identifying the optimal preprocessing technique, the complete 2D ego-motion estimation is feasible. Vertical velocity, along with roll and pitch rotations, are not estimated within this framework; it is presupposed that these components of ego-motion are pre-determined.  Thus, the 2D trajectory derived from the dataset can be visualized based on the estimated ego-motion. This trajectory is built by independently estimating ego-motion for each frame. We evaluated using relative pose error, defined as
\begin{equation}
    E_{i,j} = (P_{ref,i}^{-1}P_{ref,j})^{-1}(P_{est,i}^{-1}P_{est,j}) \in SE(2)
\end{equation}
by summing up the temporal relative error. These experiments are conducted within a planar trajectory context, rendering the relative pose error applicable in the SE(2). The trajectory is plotted in alignment with the ground truth data, utilizing Umeyama alignment method.

\subsection{Yaw Estimation Result}


During the evaluation of \ac{CFAR}, Top-k points, and Ray-max preprocessing methods, it was observed that only the \ac{CFAR} method exhibited variability in the number of feature points generated. We defined k of the Top-k points to be 200, and the Ray-max method implies the features same as the amount of rays, which is 128 in cascade radar. However, the \ac{CFAR} method demonstrated fluctuations in the quantity of feature points, directly contributing to computational bottlenecks. We verified the rapid increase in cumulative squared yaw error of \ac{CFAR} at EC hallways 0 sequence in Fig. \ref{fig:result}. Such an increase indicates the failure of the \ac{ICP} registration to converge. On the other hand, the accumulated squared error of Top-k points and Ray-max gradually increased, without any failure in both sequences. The Top-k points approach outperformed the others, and Ray-max method has a drawback where some rays omitting useful features. The \ac{RMSE} of the estimated yaw further corroborates these findings, aligning with the observed performance trends.

\begin{table}[h!]
\centering
\begin{tabular}{c|ccc}
\textbf{Dataset} & \textbf{CFAR} & \textbf{Ray max} & \textbf{Top k}  \\  \hline\hline
Aspen 5  & 1.86               &   1.94      & \textbf{1.35}               \\ 
EC hallways 0   & 4.66             & 3.09   & \textbf{2.06}                       \\ 

\end{tabular}
\caption{Yaw \ac{RMSE} [deg]}\label{tab:rmse}
\vspace{-3mm}
\end{table}

\subsection{Trajectory Result}
Given the superior performance of the Top-k preprocessing method, it was integrated with our proposed algorithm \texttt{mwICP}, which stands for the mean of the two-way weighted \ac{ICP}. Under same preprocessed features, we compared our \texttt{mwICP} method with sampled source $p_{t-1}'$ to sampled target $p_t$ weighted \ac{ICP}, and the classic weighted \ac{ICP} without any sampling steps. It was observed that the traditional weighted \ac{ICP} failed to converge to an optimal solution, as evidenced by the divergent trajectory in \figref{fig:result}. After using our sampling method, we found that the \texttt{mwICP} is more robust for the unstable feature segments of the radar point cloud. Consequently, the trajectory of the \texttt{mwICP} was notably enhanced compared to the one-way weighted \ac{ICP} especially in the EC Hallway sequence.  Through the analysis of relative pose error, it was discerned that incorporating sampling steps after preprocessing improved performance, establishing \texttt{mwICP} as the most precise method across both evaluated sequences.


\begin{table}[h!]
\centering
\begin{tabular}{c|ccc}
\textbf{Dataset} & \textbf{wICP} & \textbf{Sampling + wICP} & \textbf{Sampling + mwICP} \\  \hline\hline
EC hallways 0       & 0.0836             & 0.0124   &   \textbf{0.0086}                    \\ 
Aspen 5   &     0.0505        & 0.0077  &   \textbf{0.0066}                             \\ 
\end{tabular}
\caption{Relative Pose Error [m]}\label{tab:rpe}
\vspace{-3mm}
\end{table}

\subsection{Challenging Scenes}
\begin{figure}[!t]
   \centering
  \includegraphics[width=0.95\linewidth]{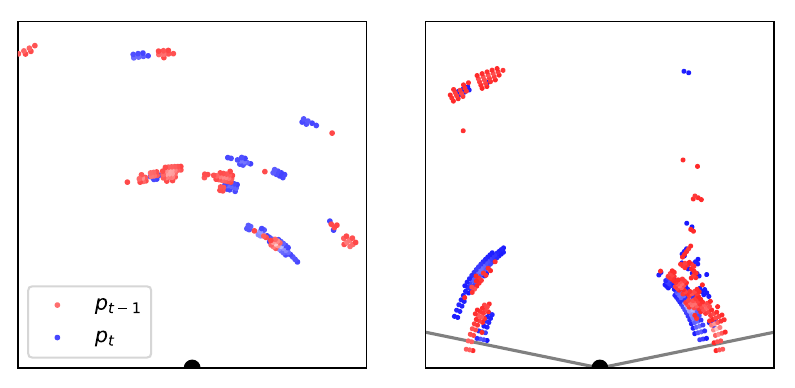}
   \caption{Challenging scenes, (left)unstable features with different segment size or (right) curvature distortion after the rectification in narrow place. Black point represent the center of the sensor and the gray lines implies the detection range of the heatmap.}
   \label{fig:fail}
\end{figure}

From the yaw error graph in \figref{fig:result}, we observed remaining errors in our method, particularly noticeable jumping errors in certain scenes. This suggests the presence of challenging scenarios for obtaining an optimal yaw rate from point registration. Drawing from our experience, we have encountered these challenging scenes in other sequences within the ColorRadar dataset, categorized into two types, as illustrated in \figref{fig:fail}.

The first scenario concerns unstable features resulting in varying segment sizes. Preprocessed points obtained through the Top-k method depend on the parameter k, leading to inconsistent segment sizes due to the noisy clutter present in radar data. This situation is particularly common when numerous small objects are present without a dominant obstacle. In attempts to mitigate this issue, we implemented an approach utilizing the mean of the two-way weighted \ac{ICP}. However, challenges persist in sequences where no suitable segment target is identified in successive frames.

The second scenario involves the curvature distortion of close points after rectification. This phenomenon is frequently observed in narrow hallways where preprocessed points densely populate the close area. After rectifying the previous frame, the coordinate system aligns with the current frame, resulting in a change in the center of rotation. As feature points typically manifest as a circular sector within a certain range, distortion during rectification worsens when feature points are in close proximity. Additionally, as the azimuth resolution increases towards both sides, this effect contributes to additional errors in narrow scenes.

%% file: conclusion.tex
\section{Conclusion}
\label{sec:conclusion}
In this work, we present an approach for estimating 2D ego-motion, including yaw rate, by matching feature points derived from mmWave radar heatmap data. The accuracy of point cloud registration was further enhanced by adopting the weighted \ac{ICP} algorithm, particularly via a bidirectional methodology that accommodates both forward and reverse computations. The proposed method's validity was substantiated by depicting trajectories derived from estimated ego-motion. While the current study focuses on the heatmap generated by cascade radar systems, future endeavors aim to extend the application of this methodology to heatmap data obtained from single-chip systems.